\documentclass{llncs}
\usepackage{graphicx}
\usepackage{amsmath}
\usepackage{amssymb}
\usepackage{tikz}
\usepackage{algorithm}
\usepackage{tabularx}
\usepackage{gensymb}
\usepackage{algpseudocode}
\usepackage{enumerate}
\newcommand{\mypm}{\hspace{0.06cm}$\pm$\hspace{0.06cm}}
\begin{document}

\title{Extreme Points Derived Confidence Map as a Cue For Class-Agnostic Segmentation Using Deep Neural Network}

\author{Shadab Khan\thanks{= equal contribution} \and Ahmed H. Shahin* \and Javier Villafruela \and Jianbing Shen \and Ling Shao}
\institute{Inception Institute of Artificial Intelligence, Al Bustan Offices, Abu Dhabi, UAE\\
\email{skhan[dot]shadab[at]gmail}\\
}

\maketitle

\begin{abstract}
To automate the process of segmenting an anatomy of interest, we can learn a model from previously annotated data. The learning-based approach uses annotations to train a model that tries to emulate the expert labeling on a new data set. While tremendous progress has been made using such approaches, labeling of medical images remains a time-consuming and expensive task. In this paper, we evaluate the utility of extreme points in learning to segment. Specifically, we propose a novel approach to compute a confidence map from extreme points that quantitatively encodes the priors derived from extreme points. We use the confidence map as a cue to train a deep neural network based on ResNet-101 and PSP module to develop a class-agnostic segmentation model that outperforms state-of-the-art method that employs extreme points as a cue. Further, we evaluate a realistic use-case by using our model to generate training data for supervised learning (U-Net) and observed that U-Net performs comparably when trained with either the generated data or the ground truth data. These findings suggest that models trained using cues can be used to generate reliable training data.
\end{abstract}

\section{Introduction}

Deep neural networks have enabled tremendous progress in medical image segmentation. This progress has been greatly enabled by the large quantity of annotated data. Supervised techniques trained with large annotated data, have accomplished outstanding results on many segmentation tasks. However, the annotations need to cover the inter- and intra-patient variability, tissue heterogeneity, as well as lack of consistency between imaging scanners, operators, and annotators. As a result, image labeling is slow, expensive, and subject to availability of annotation experts (clinicians), which varies widely across the world.

\begin{figure}[t]
\centering
\vspace{-3mm}
\includegraphics[scale=0.462]{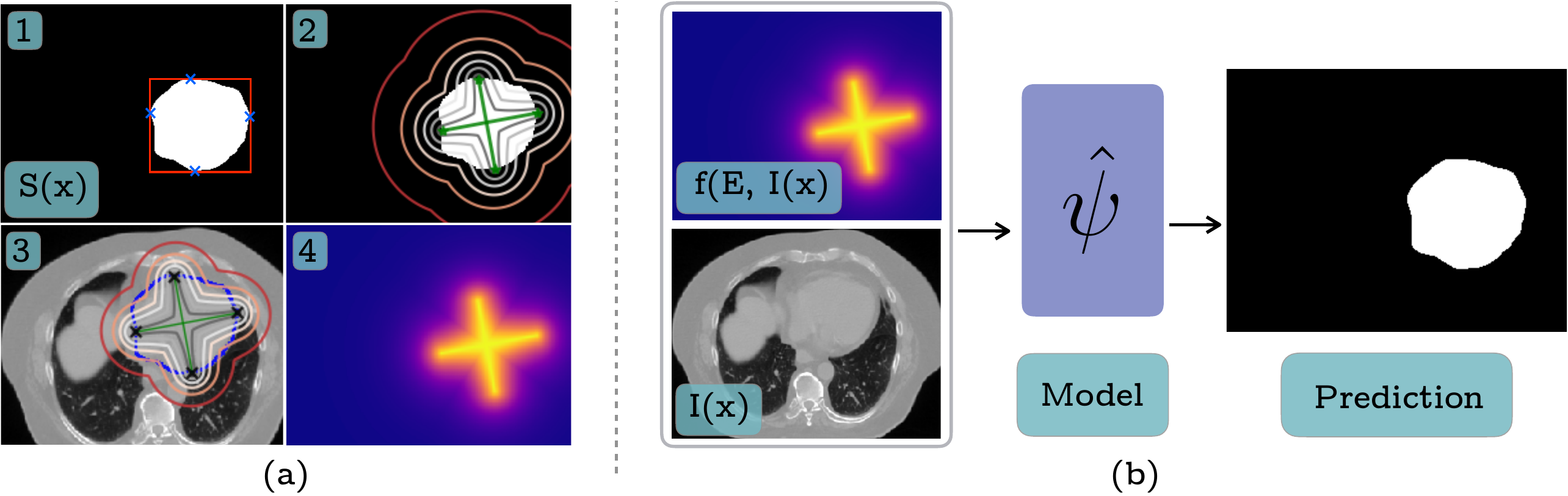}
\vspace{0mm}
\caption{(a) Extreme points (`x') and bounding box(red) shown on a given segmentation $S(x)$ (1), which was used to compute the confidence map (CM) $f(E,I(x)$ (4) as shown. The CM produces iso-contours with negative curvature, this is desired as explained in section 2.1. (3) shows iso-contours overlaid on the image, where boundary of segmented region has been shown in blue and user-clicked extreme points as `x' markers. (b) For inference, a confidence map computed using extreme points is input to the model.} 
\label{fig:fig1}
\vspace{-2mm}
\end{figure}

To address this issue, techniques that can employ cues such as image label, scribbles, bounding box, and more recently, extreme points, (fig.~\ref{fig:fig1}) have been used to enable weakly supervised training with results that are comparable to those obtained using ground truth pixel-level segmentation ~\cite{Cai2018,Maninis2018,Papadopoulos2017,Rajchl2017,Schlegl2015}. Learnt models that produce pixel-level segmentation from user-provided cues can then be used to significantly accelerate the annotation process. This approach has the advantage of exploiting existing annotations as a prior knowledge to enable the annotation of a new related data set. In this paper, we evaluate the utility of extreme points as a cue for medical image segmentation. Extreme points can be labeled more quickly than bounding boxes ($\sim$7.2 seconds vs $\sim$34.5 seconds), as shown by a recent study~\cite{Papadopoulos2017}, and implicitly provide more information to the learning models as they lie on the object of interest.

We explore a novel algorithm to encode information from extreme points and generate a confidence map to guide the neural networks in understanding where the object lies within the extremities defined by the extreme points. The training data is augmented with confidence map to train a model that produces accurate segmentation, using confidence map and image as input. Further, we present an algorithm for fast computation of distance of points from a line segment that allows us to generate confidence maps during training and keeps the memory footprint low. We tested our approach against the state-of-the-art method in employing extreme points as a cue~\cite{Maninis2018} under identical and unbiased settings and found that our approach improves the segmentation performance for all organ categories in the multi-class SegTHOR data set~\cite{Trullo2017}. We also evaluated the algorithm under a use-case scenario (labeling a new data set) and found that supervised training using segmentation produced by our approach performs well compared to when the ground truth segmentation were used for training.

\section{Methods}

\subsection{Problem Formulation}
Given an image $I(x) \in \Omega; x \in \mathbb{R}^2$ and the four extreme points $E= {x_1,x_2,x_3,x_4};$ $x_i \in \mathbb{R}^2$, we aim to compute a segmentation $S(x)$ of the image, such that $S:\Omega\mapsto\{ 0,1 \}$. This is accomplished using a segmentation map $\mathcal{\psi}:\Omega\mapsto\Omega$. In a supervised learning setting, an approximate map $(\hat{\mathcal{\psi}})$ is learnt using a set of training pairs $\{(I(x), S(x))_i;i\in[1,N]\}$, with cardinality $N$. In our approach, we propose to learn $(\hat{\psi})$ using an augmented training set $\{(f(E,x),I(x), S(x))_i\}$, where $f(E,x):\Omega\mapsto\Omega$ is a function that assigns a confidence score to every point in the image domain. Our objective is to develop a class-agnostic $(\hat{\psi})$ that can segment a region-of-interest using the points in $E$ as a cue.

To accomplish this, we propose to exploit the following cues: (i) The extreme points form line segments $\overline{x_1x_2}$ and $\overline{x_3x_4}$ respectively that have a point of intersection, denoted as $c$, (ii) $\overline{x_1x_2}$, $\overline{x_3x_4}$, and $c$ are likely to lie on region-of-interest (RoI), (iii) Points in $\Omega$ away from $c$, $\overline{x_1x_2}$ or $\overline{x_3x_4}$ are less likely to lie on RoI, (iv) Points lying on $\overline{x_1x_2}$ or $\overline{x_3x_4}$ are likely to lie on the RoI. We formulate $f(E,x)$ to take into account this prior information while assigning a RoI-membership confidence score to each point in $\Omega$.

Before explaining our generalized formulation for $f(E,x)$, we consider a simpler case where two assumptions are made: (i)$\overline{x_1x_2}\perp\overline{x_3x_4}$, (ii) $c$ bisects $\overline{x_1x_2}$, and $\overline{x_1x_2}$. In this scenario, assuming that the lengths of $\overline{x_1x_2}$ and $\overline{x_3x_4}$ can be used to approximate the measure of spread of the RoI (variance along $\overline{x_1x_2}$ and $\overline{x_3x_4}$), the following formulae allow us to incorporate the priors with one exception (explained ahead):

\begin{equation}
\label{eq:chebyshev1}
d_1(x) = min\odot \{ R^{-1}(x-c)\Lambda ^{-\tfrac{1}{2}} \}, \hspace{6mm}
d_2(x) = \{ (x-c)^TS^{-1}(x-c) \}^{\tfrac{1}{2}}
\end{equation}

\begin{equation}
\label{eq:weight1}
d_3(x) = \begin{cases}
 & 1 \text{ if } d_2(x)\leqslant \tau  \\ 
 & 0 \text{ if } d_2(x)> \tau
\end{cases}, \hspace{6mm}
f(E,x)=\frac{d_3(x)}{1+d_1(x)d_2(x)}
\end{equation}
In the equations above, $min\odot$ is an element-wise minimum taken over the resulting vector, $S$ is the covariance matrix of the data (foreground pixels in the $S(x)$), and $R$ and $\Lambda$ are obtained by decomposing $S$ as: $S=R\Lambda R^T$, where $R$ represents the rotation matrix that rotates the standard axes into alignment with $\overline{x_1x_2}$ and $\overline{x_3x_4}$, $\Lambda$ is the diagonalized covariance matrix, and $\tau$ is a threshold. In equation \eqref{eq:chebyshev1}, $d_1(x)$ measures an equivalent of Chebyshev distance, and $d_2(x)$ measures the Mahalonobis distance in the coordinate frame of $\overline{x_1x_2}$ and $\overline{x_3x_4}$. Fig. 1 shows $d_1(x)$, $d_2(x)$, $d_3(x)$, and $z(x)$. This formulation places an equal weight along the line $\overleftrightarrow{x_1x_2}$, which is a departure from the priors.

To overcome this limitation, we use the following formulae:
\begin{equation}
\label{eq:generalized_xy}
d_{\overline{x_1x_2}}(x) = \frac{dist(x,\overline{x_1x_2})}{\sigma_{\overline{x_1x_2}}}, \hspace{6mm}
d_{\overline{x_3x_4}}(x) = \frac{dist(x,\overline{x_3x_4})}{\sigma_{\overline{x_3x_4}}}
\end{equation}
\vspace{-3mm}
\begin{equation}
\label{eq:generalized_chebyshev}
\hat{d_1}(x) = min \{d_{\overline{x_1x_2}}(x),d_{\overline{x_3x_4}}(x) \}, \hspace{2mm}
\hat{d_2}(x) = \{ d_{\overline{x_1x_2}}(x)^2+d_{\overline{x_3x_4}}(x)^2 \}^{\tfrac{1}{2}}
\end{equation}
\begin{equation}
    \label{eq:generalized_weight}
    f(E,x) = \frac{1}{1+\hat{d}_1(x)\hat{d}_2(x)}
\end{equation}
where $dist(x,\overline{x_1x_2})$ is the distance of point $x$ from line segment $\overline{x_1x_2}$ and $\sigma_{\overline{x_1x_2}}$ approximates the variance along ${\overline{x_1x_2}}$. Equation \eqref{eq:chebyshev1} is a special case of eq. \eqref{eq:generalized_chebyshev}, when $x_1$ and $x_2$ are at $\pm\infty$, $x_3$ and $x_4$ are at $\pm\infty$, and $\overline{x_1x_2}\perp \overline{x_3x_4}$. By including $f(E,x)$ with $I(x)$ and $S(x)$ we create an augmented data-set that is used for computing $(\hat{\psi})$.

\section{Implementation}
\subsection{Model and Data Set}
We use a deep neural network with ResNet-101 architecture~\cite{He2016} to approximate $(\hat{\psi})$, with a few changes. The fully-connected layers and the final two max-pool layers at the end of the ResNet-101 architecture are removed and atrous convolution is added in the final two layers. Lastly, a Pyramid Scene Parsing (PSP)~\cite{Zhao2017} module is incorporated at the last stage to introduce global context. To experiment with medical images where multiple organs have been annotated, we chose SegTHOR data set ~\cite{Trullo2017}. SegTHOR data set comprises annotated CT images of heart, aorta, trachea and esophagus. The soft tissue in heart, aorta and esophagus have a closely matching dynamic range in Hounsfield Units(HU) and therefore present challenging conditions for testing segmentation performance. 

\subsection{Data Pre-processing and Model Setting}
SegTHOR comprises CT scans of 40 patients, acquired with 0.9--1.37mm in-plane ($512\times512$ field-of-view) and 2--3.7mm out-of-plane resolution resulting in 150--284 slices per patient. Heart, trachea, esophagus, and aorta were annotated in a total of 7390 slices. To create our training data, 4 extreme points were deduced for each organ in all annotated slices using ground truth segmentation. The input to the neural network was a resized crop of the anatomy with dimensions $512\times512$. To create the input to the neural network, a bounding box of dimensions $w\times h$ was calculated using the extreme points. Next, using $b=max(w,h)$, we calculated a zoom factor $z$, such that $z= b_{m}/b$, where $b_{m}$ is a random number in [350,400]. This approach of calculating $z$ ensures that approximately 45--60$\%$ pixels seen by the network belong to the anatomy of interest. Images were windowed (-200, 250) and intensity normalized before input to the network.

We used the implementation of ResNet-101 and PSP module provided by~\cite{Chen2018} and~\cite{Maninis2018}. The network was initialized using pre-trained weights for a 4-channel version provided by~\cite{Maninis2018}. We fine-tuned the network using a learning rate of 1e-7, batch size=14, Adam optimizer~\cite{kingma2014adam} ($\beta_{1}=0.90$, $\beta_{2}=0.99$), L2-regularization ($\alpha$=5e-4) and loss function set to weighted cross entropy. Data augmentation in the form of random scaling (0.9--1.1), rotation ($-30\degree$ to $+30\degree$), and horizontal flip was used. Data was split at patient-level into 60/20/20 splits for training, validation, and test respectively. Training loop was executed for 100 epochs, and model selection was done by evaluating validation set performance. To report results, the best model was tested on test set only once.

\subsection{Confidence Map ($f(E,x)$)}
\begin{figure}[t]
\centering
\vspace{-3mm}
\includegraphics[scale=0.462]{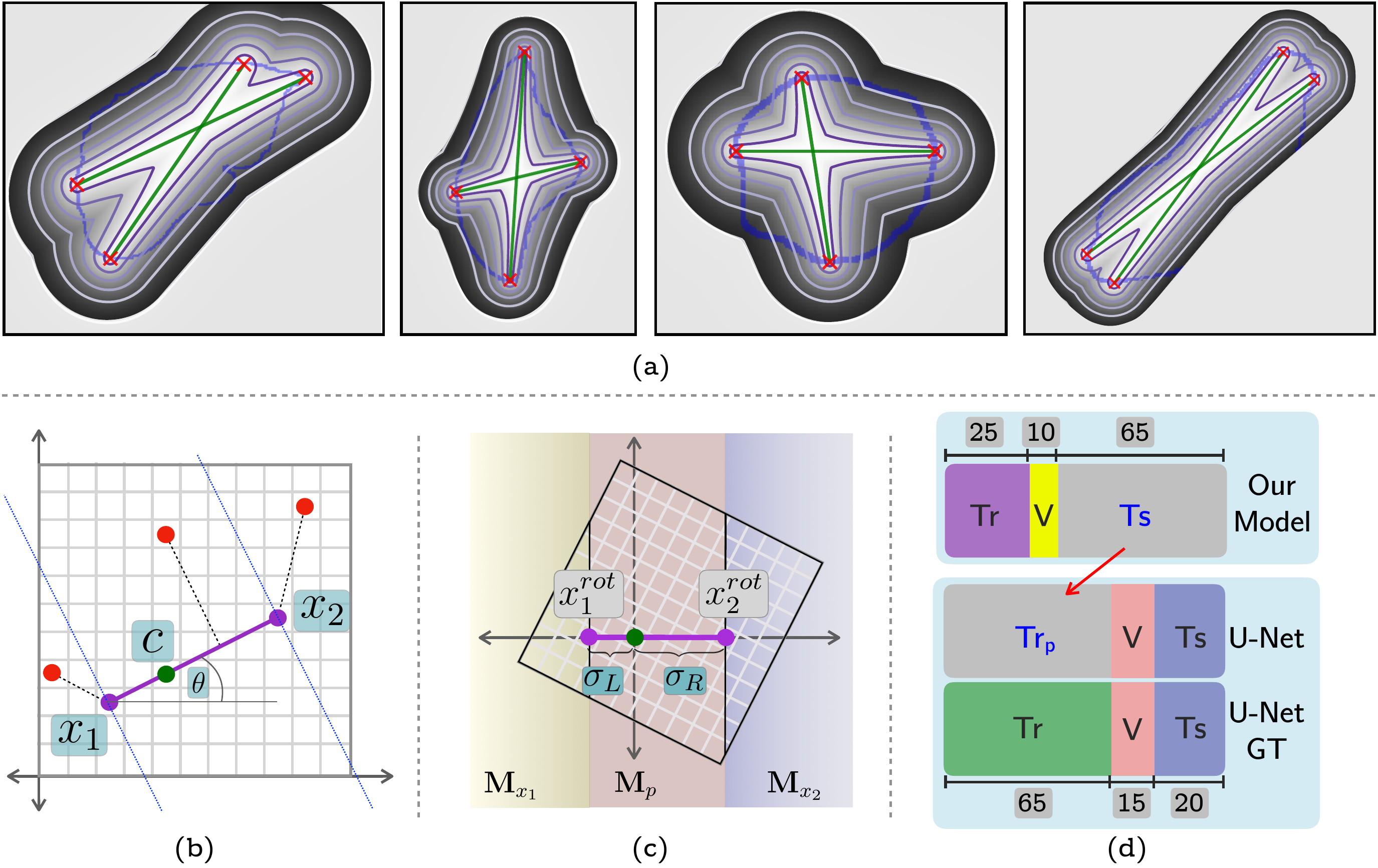}
\vspace{0mm}
\caption{(a) Iso-contour plots are shown overlaid on the computed confidence map for a few different object shapes. For visualization, we set the area outside last contour to a gray intensity, blue line shows the boundary of the object, and red `x' markers are the extreme points. We draw attention to how the iso-contour lines flex and bend to retain the negative curvature characteristic across varying object shapes. (b) and (c) help explain Algorithm 1, and (d) shows how the CHAOS data was split for reporting results in section 4.2, where Tr=Train, V=Val, Ts=Test, and numbers (`65', `25', etc) show the \%. `U-Net GT' was trained using ground truth segmentation, and `U-Net' was trained using segmentation produced by our model.} 
\label{fig:fig2}
\vspace{0mm}
\end{figure}

Computing $f(E,x)$ requires evaluating distance of each point in $\Omega$ from the line segments $\overline{x_1x_2}$ and $\overline{x_3x_4}$. This is non-trivial if $\Omega$ is large. A time-efficient solution was obtained by implementing calculation of distance from line segments for all points in $\Omega$ as follows (Fig.~\ref{fig:fig2}):

\vspace{-2mm}
  \begin{algorithm}
   \caption{Compute distance of points in image from line segment $(\mathbf{D}_{\overline{x_1x_2}})$}
    \begin{algorithmic}[1]
        \State Calculate $\text x_{-}$, $\text x_{+}$, and $\text y_{-}$, $\text y_{+}$ as the extent of image size
        \State Create 2D arrays $\mathbf X$ and $\mathbf Y$ as: $\mathbf X$, $\mathbf Y$ $\leftarrow$ $meshgrid$($\text x_{-}$, $\text x_{+}$, $\text y_{-}$, $\text y_{+}$)
        \State $\mathbf X=\mathbf X-c$; $\mathbf Y = \mathbf Y-c$
        \State $x_1 = x_1-c$; $x_2 = x_2-c$
        \State Calculate unit vector along ${\overline{x_1x_2}}$ as: $cos(\theta)\hat{i}+sin(\theta)\hat{j}$\vspace{1mm}
        \State $\mathbf X^{rot} \leftarrow \mathbf X cos(\theta) - \mathbf Y sin(\theta)$\textbf{;} $\hspace{5mm}\mathbf Y^{rot} \leftarrow \mathbf X sin(\theta) + \mathbf Ycos(\theta)$\vspace{1mm}
        \State $x^{rot}_{1i}\leftarrow x_{1i} cos(\theta) - x_{1j} sin(\theta)$\textbf{;} $\hspace{3mm} x^{rot}_{1j}\leftarrow 0\textbf{;}  \hspace{5mm} x^{rot}_{1}$ lies along $1\hat{i}+0\hat{j}$\vspace{1mm}
        \State $x^{rot}_{1} = (x^{rot}_{1i},0)$; similarly calculate $x^{rot}_{2}$\vspace{1mm}
        \State $\mathbf{D}_{x_1} = \sqrt{(\mathbf X^{rot}-x^{rot}_{1i})^2+(\mathbf Y^{rot})^2}$; similarly calculate $\mathbf{D}_{x_2}$\vspace{1mm}
        \State $\mathbf{D}_{p}=\left | \mathbf Y^{rot} \right |$\vspace{1mm}
        \State $\mathbf M_{x_1}=\mathbf X^{rot} > x^{rot}_{1i}\textbf{;}\hspace{5mm} \mathbf M_{x_2} = \mathbf X^{rot} < x^{rot}_{2i}\textbf{;}\hspace{5mm} \mathbf{M}_{p}=\neg\mathbf M_{x_1} \land \neg\mathbf M_{x_2}$
        \State $\mathbf{D}_{\overline{x_1x_2}}=\mathbf M_{x_1}\mathbf{D}_{x_1}+\mathbf M_{x_2}\mathbf{D}_{x_2}+\mathbf{M}_{p}\mathbf{D}_{p}$\vspace{1mm}
        \State $\mathbf M_R = \mathbf X^{rot} > 0\textbf{;}\hspace{5mm} \mathbf M_L = \mathbf X^{rot} \leq 0$ \hspace{10mm} \hfill \# Right and left mask
        \State $\sigma_L= \left | x_{1i}^{rot} \right |\textbf{;}\hspace{5mm} \sigma_R= \left | x_{2i}^{rot} \right |$ \hspace{5mm} \hfill \# Approximation to right and left variance
        \State $\mathbf \Sigma = \sigma_R\mathbf M_R + \sigma_L\mathbf M_L $
        \State $\mathbf{D}_{\overline{x_1x_2}}=\mathbf{D}_{\overline{x_1x_2}}/\mathbf \Sigma$
        
 

    \end{algorithmic}
\end{algorithm}
\vspace{-5mm}

Above, $meshgrid()$ is a computer program, boldface letters are 2D arrays, $rot$ refers to `rotated', and all $\mathbf M$'s are 2D boolean arrays. This algorithm can be implemented without using any loops in Python and can be used to generate confidence maps during training itself. On a CPU equipped with 2.2 GHz Intel Xeon 5120 processor, it took 88 milliseconds to compute $\mathbf{D}_{\overline{x_1x_2}}$ for an image size $512\times512$. Fig.~\ref{fig:fig2} helps explain the algorithm. The confidence map $f(E,x)$ was incorporated into the input as an extra channel passed to the neural network.

\section{Results}
\subsection{Our Approach and Baseline}
In testing the performance of our approach, our objective was to evaluate how to best encode information from the 4 extreme points for class-agnostic segmentation. We evaluated our model's performance on unseen data by evaluating mean Dice overlap score on $\sim$20\% of the patients from SegTHOR, the model was not exposed to any slice from this set at any time during training or validation. For baseline comparison, we also fine-tuned a state-of-the-art pre-trained model~\cite{Maninis2018} with hyper-parameter, data pre-processing steps, and all other settings identical to the one used to test our model. The baseline model places Gaussians at the extreme points and was shown to outperform all other methods that employ cues for segmentation (including GrabCut (GC)~\cite{Rother2004}). In our experiments, we initialized GC by setting area outside bounding box extended by extreme points to background, and extreme points to foreground, however GC did not produce any meaningful segmentation to warrant further exploration. We also evaluated combination of the confidence map and extreme points (gaussians) as two separate input channels. The mean Dice score results for all experiments are summarized in table 1, organized by testing on each organ.

\begin{table}[b]
\begin{center}
\label{my-label}
\newcolumntype{C}{>{\centering\arraybackslash}m}
\setlength\extrarowheight{2pt}
\begin{tabular}{C {0.16\textwidth}
                C {0.2\textwidth}
                C {0.2\textwidth}
                C {0.2\textwidth}
                C {0.2\textwidth}}
\hline
 Organ & CM (\textbf{Ours}) & EP & CM+EP (\textbf{Ours}) & Best (No Cues) \\ \hline
Aorta & 94.00\mypm2.02 & 92.80\mypm1.89& \textbf{94.41}\mypm\textbf{1.87} & 86\mypm5  \\ 
Esophagus & \textbf{89.87}\mypm4.36 & 88.14\mypm4.50 & 89.83\mypm\textbf{4.16} & 67\mypm 4\\ 
Heart & 95.97\mypm2.09 & 95.41\mypm2.05 & \textbf{96.53}\mypm \textbf{1.94} & 90\mypm 1\\ 
Trachea & \textbf{91.87}\mypm4.07 & 90.05\mypm\textbf{3.90} & 91.24\mypm 4.27 & 83\mypm 6\\ \hline
\end{tabular}
\end{center}
\caption{Mean Dice (mDice) score (\%) on SegTHOR data organized by organ type. CM--Confidence Map, EP--Extreme Points. CM achieves higher mDice compared to EP. CM+EP inputs CM and EP (gaussians) as separate input channels. Results in the last column are the best achieved by a fully-supervised model without using any cue.}
\label{tab:table}
\end{table}

\subsection{Weakly Supervised Segmentation}
In order to evaluate the efficacy of our model in producing accurate segmentation for fully supervised training, we fine-tuned our model on the segmentation data of a new-to-the-model organ (liver, CHAOS data set~\cite{Selver2018}). Using a patient-level 25/10/65 split of the data (fig.~\ref{fig:fig2}d), we produced segmentation on data collected from 65\% of the patients for which ground truth segmentation was also available. Next, we trained 2 versions of U-Net~\cite{Ronneberger2015} in a fully-supervised manner using a patient-level 65/15/20 split of the data, as shown in fig.~\ref{fig:fig2}. Both versions of the U-Net were identical in experimental settings (hyper-parameters, training, starting model) and training images used, and differed by using either ground-truth segmentation (U-Net GT) or the ones generated by our model (U-Net) for training. The U-Net GT model achieved mean Dice score (\%) of 91.70\mypm13.00, compared to 90.35\mypm9.89 by U-Net trained on generated segmentation.

\section{Discussion and Conclusion}
We evaluated a new approach to incorporate cues from extreme points into neural network by computing a confidence map that is used during training. This was enabled by our algorithm for quickly computing distance of points from line segment. Our approach, when compared with the state-of-the-art baseline under identical and unbiased conditions resulted in improved mean Dice score across all four organs in the test set, with closely-matching variance in the Dice scores across samples. Interestingly, a combination of our confidence map and extreme points (gaussians) further improved the mean Dice for 2 out 4 organs while reducing variance. This strongly suggests that confidence map provide superior guidance to neural networks for segmentation, compared to extreme points alone.

On qualitative evaluation, the segmentation results were found to be consistent. We probed the samples which resulted in lower Dice score compared to the group mean and observed that these were slices where the organ occupied a small area within the image. Resizing such instances for input to the network is associated with two factors: (i) lack of texture in resized image, and (ii) rescaling binary segmentation can introduce non-trivial noise in the loss function. We posit these factors reduce the quantitative measures of segmentation performance. We further evaluated our model's ability to produce segmentation for fully-supervised learning. We observed that U-Net trained using segmentation produced by our model achieves slightly lower mean Dice score than the gold standard (U-Net GT), but achieves lower variance compared to U-Net GT. 

Our findings suggest that to quickly annotate large data sets, it may suffice to: 1) Fine-tune a pre-trained model using a fully annotated small proportion of the data, 2) Use pre-trained model along with extreme points as cue to predict segmentation on the rest of the unlabeled data, 3) Use the generated labels to train a fully-supervised algorithm. Such an approach would help reduce the annotation time and expense drastically and allow more data to be labeled.

\bibliographystyle{splncs03}
\bibliography{MICCAI_19}

\end{document}